\documentclass[11pt]{article}
\usepackage[utf8]{inputenc}
\usepackage[T1]{fontenc}
\usepackage[margin=1in]{geometry}
\usepackage{authblk}
\usepackage[backend=biber, style=authoryear, maxcitenames=2, uniquelist=false]{biblatex}
\usepackage{hyperref}
\usepackage{graphicx}
\usepackage{float}
\usepackage{subcaption}
\usepackage{booktabs}
\usepackage{tabularx}
\newcommand{\keywords}[1]{\begin{center}\textbf{#1}\end{center}}

\title{Frankly Speaking: The Company Election Integrity Keeps in Congressional Communication}

\author[1,*]{Michael A. Robinette}
\author[3]{Christina Barsky}
\author[3]{Amanda D. Clark}
\author[1]{Helen Flannery}
\author[1,2,4]{Juniper Lovato}

\affil[1]{Vermont Complex Systems Institute, University of Vermont, Burlington, 05405, U.S.A.}
\affil[2]{Department of Computer Science, University of Vermont, Burlington, 05405, U.S.A.}
\affil[3]{Community Development and Applied Economics, University of Vermont, Burlington, 05405, U.S.A.}
\affil[4]{Complexity Science Hub, Vienna, Austria}

\affil[*]{Michael.Robinette@uvm.edu}
\begin{document}

\flushbottom
\maketitle

\keywords{Keywords: election integrity $|$ elite rhetoric $|$ partisan communication $|$ congressional newsletters $|$ diachronic word embeddings} 

\begin{abstract}
Election integrity is a shared commitment in American politics, but Democrats and Republicans in Congress increasingly mean different things by it. Analyzing 10,420 congressional e-newsletters sent from 2010 to 2025, we examine how the two parties invoke election terms such as ``integrity,'' ``security,'' and ``fraud.'' Using diachronic word embeddings, which track semantic change through how words are used in context, we measure how differently each party uses each term, how much each party shifts, and whether they share nearest semantic neighbors. We show that members of Congress do not need different words to tell different stories about elections. They need only place those same words in different company. Across nearly every term and year, the two parties surround election-integrity vocabulary with entirely different words. That divergence is easy to miss, because most election vocabulary is no more divergent than other words used at roughly the same frequency, and neither party's usage shifted more than the other's. Only a subset of terms grew more divergent, most notably ``certification,'' while ``rigged'' and ``stolen'' entered election discourse after 2020 and polarized there. By 2024, Democrats surround ``voter,'' ``election,'' and ``integrity'' with the language of access and administration. Republicans surround the same terms with ``citizenship verification,'' ``noncitizen voting,'' and ``border security.'' For one party, integrity means protecting eligible voters' ability to cast a ballot. For the other, it means protecting the electoral process from illegitimate ballots. Election officials, who must administer a single system under both definitions, inherit this tension.
\end{abstract}

\section{Introduction}
\label{sec:introduction}

Election integrity has long been recognized as a cornerstone of democratic governance, but in recent years the term has taken on sharper political edges. In both legislative debate and public communication, words like ``integrity,'' ``security,'' and ``fraud'' now often appear in charged partisan contexts. While prior scholarship has documented changes in voter confidence and partisan polarization, there has been less attention to how integrity-related rhetoric is constructed, deployed, and reinterpreted over time. Existing work tends to focus either on legal standards and administrative performance, or on mass beliefs about ``rigged'' elections and fraud, leaving a gap in our understanding of how elites themselves narrate and redefine election integrity over time. Election integrity is a powerful symbolic indicator of partisan commitments. 
Email newsletters are a primary tool that elected representatives use to communicate with their constituents
 \autocite{blum2023conditional, cormack2016extremity, grimmer2010bayesian, grimmer2013text, mathur2023manipulative}.  Newsletters discuss activities, policy positions, and policy actions of the elected representative. They provide a window into how elected officials communicate with constituents outside traditional, filtered media \autocite{gamson1989media, graf2015image, tripodi2022you}. Because these messages are official, publicly funded communications (rather than fundraising or campaign appeals), they occupy a distinct space at the intersection of public administration, representation, and partisan messaging, making them a useful lens for examining how members of Congress talk about election integrity in their governing.

We draw on a longitudinal corpus of congressional e-newsletters from 2010-2025 and employ natural language processing (NLP) techniques to identify semantic changes and partisan divergence in meaning surrounding election integrity. Using sixteen initial target election integrity terms, we measure how differently the two parties use each term, how much each party’s usage shifts over time, and which words each party uses in the company of those terms.This is a term’s semantic neighborhood: its nearest neighbors in a word embedding model. In Firth’s \autocite{firth1957synopsis} phrase, a term’s meaning comes from the company it keeps (its semantic neighborhood). 

This study situates the rhetorical patterns found in these newsletters within the broader contexts of election law, public administration, and political behavior, highlighting the potential consequences for institutional legitimacy and voter confidence. Identifying whether the language of integrity is converging, diverging, or fragmenting along partisan lines can illuminate how trust in electoral processes is reinforced or eroded through elite discourse. 

We see that while partisan distance and within-party shift measures show modest results, some terms, notably ``certification,'' diverged more over time. The clearest differences lie in the contextual semantic networks that contain these terms. By 2024, Democrats embed terms like ``voter,'' ``election,'' and “integrity” in networks of access and administration, while Republicans embed the same terms in networks of citizenship verification and border security. These findings contribute to theories of symbolic politics by showing how a term like ``election integrity,'' which neither party opposes, can carry partisan meanings through the company it keeps. They also carry a practical lesson for anyone trying to hold a bipartisan conversation about elections, from election officials to reform commissions. Shared vocabulary is not necessarily shared meaning, and discourse about electoral legitimacy can only be meaningfully understood once we know how each side is using these terms in context.

\section{Theoretical Framework and Literature Review}
\label{sec:theory}

\subsection{Election integrity, fraud, and trust}

Election integrity is not a singular, uncontested concept; it should be understood as a cluster of overlapping concerns about accuracy, security, access, and equal treatment  \autocite{Barsky2026}. Much legal and technical scholarship considers election integrity to be concerned primarily with correctness, focusing on verifiable outcomes and preserving ballot secrecy \autocite{bernhard2017public, kuriwaki2025privacy}. By contrast, public and election administration perspectives center integrity around fairness, including the accessibility and equality of the process \autocite{james2021delivering, porter2018partisanship}. Together these views create two competing conceptions: integrity as protection against illegitimate votes, and integrity as protection of access for eligible voters.

In such a contested conceptual space, elite rhetoric about fraud, security, and accuracy plays an important role in shaping how the public understands election integrity. Debates about voter identification, registration purges, and convenience voting methods have become coupled with escalating claims related to “voter fraud,” “rigged elections,” and the need to “secure” the vote, despite repeated empirical findings that fraud in American elections is rare. Previous research highlights how elite claims of fraud reduce public confidence in electoral integrity \autocite{berlinski2023effects}, strengthen beliefs that elections are ``rigged'' \autocite{clayton2021elite}, reduce perceptions of fairness \autocite{justwan2022trump}, and decrease confidence in administrative procedures \autocite{mongrain2023suspicious}.

Members of Congress (MOCs) occupy a distinct position in this environment as lawmakers, party members, and highly visible communicators. Explorations of legislative communication show that MOCs strategically tailor their messages to signal priorities, claim credit, and assert partisan positions on controversies \autocite{cormack2016extremity, fenno1978home, green2024cross, mathur2023manipulative}. Studies of MOC partisan positioning across communication modes further demonstrate that elite rhetoric often is more polarized when shared externally and via partisan-friendly channels than in institutional settings \autocite{green2024cross}, suggesting MOCs use these venues to mobilize others around issues like election integrity. How MOCs talk about election integrity is not incidental: it is part of an elite repertoire that can prime others to view the term as indicative of fraud and manipulation or of fairness and competence.

\subsection{Symbolic politics, elite communication, and problem construction}

Because integrity can mean either protection from fraud or protection of rights, elites can deploy it to signal different priorities without relinquishing the high ground. Classic work on symbolic politics \autocite{edelman1964symbolic} argues that officials regularly deploy emotionally resonant but ambiguous symbols to create the appearance of consensus, mobilize support, and mask underlying disruptions or conflicts. In these terms, election integrity functions as a resonant but deliberately under-specified symbol through which MOCs can project vigilance and democratic commitment while obscuring the distributive and administrative tradeoffs in concrete policy actions.

Symbolic issues are particularly attractive when the underlying problems are complex, intractable, or politically costly to address because it allows officials to do something visible to constituents while leaving core structures unchanged \autocite{edelman1964symbolic}. This suggests that MOCs may take up the concept of election integrity precisely because the root challenges---federalism, resource disparities, and party incentives around turnout, and others---are difficult to resolve. Treating election integrity as a symbolic talisman of vigilance, rather than as a concrete standard, enables MOCs to claim credit for ``protecting our elections'' even when our elections are already free and fair \autocite{brennan2022myth, jouvenal2023gop, wolfe2022debunked}.

Symbolic issues require a careful construction of a social or political problem that can capture the public's attention and make its way into policy \autocite{kingdon1984agendas}. Causal stories, or narratives, are integral to defining political problems and creating shared identities \autocite{stone1989causal}. The ability for people to recognize their in-group through coded language and problem frames mobilizes them to act \autocite{benford2000framing, snow1986frame}. The First Continental Congress granted members the `franking' privilege of free constituent mail in 1775, and that channel has been protected and regulated ever since \autocite{hadijski2020modernizing}. While all MOCs continue to use traditional modes of communication, changes in technology allow elected officials to also reach their constituents via email and other electronic means \autocite{straus2016social}. Congressional newsletters are seen as crucial for accountability, agenda-setting, and legitimacy across a geographically large and information-unequal democracy. MOCs have turned increasingly to electronic newsletters (``e-newsletters'') as low-cost, low-barrier opportunities to reach constituents and manage their public image \autocite{davis2025redefining}; these are analogous to the physical newsletters lawmakers traditionally sent via franked mail \autocite{hunt2025modern}.

Communications sent from an MOC in their official capacity, like e-newsletters, cannot legally solicit campaign contributions or make explicit appeals for support, and instead are expressly intended to provide information to constituents about their representative's activities. Compared to other tools (social media posts, press releases, one-minute floor speeches), e-newsletters are ``structured around home state and district concerns and are primarily meant for an audience of in-district constituents who have opted in to receive information from that specific member'' \autocite[656]{green2024cross}. E-newsletters have supported a growing literature on legislative effectiveness, identity, rhetoric, and strategy in the U.S.\ and abroad \autocite{bach2025let, casey2025canberrainbox, huffman2025proving, hunt2025modern, jackson2006mp, kraft2024lexical, marland2023representation, schmitt2023conspiracy, umit2017happiness}.

\subsection{Text-as-data and computational analysis of political language}

The text-as-data literature treats bills, speeches, and press releases as systematic evidence of elite behavior \autocite{grimmer2022text}, and computational methods have uncovered patterns difficult to see otherwise \autocite{bisbee2025yellin, gaynor2025following, grimmer2013text, hughes2022party, matalon2021using, meier2026toward}.

If a political symbol's meaning lives in the narratives elites attach to it, then that meaning should show up in the words that regularly surround it. This intuition comes from philosophy and linguistics: \textcite{wittgenstein1968} argued that in many cases the meaning of a word is its use in the language, and \textcite{firth1957synopsis} made the parallel claim for empirical linguistics---that you shall know a word by the company it keeps. \textcite{harris1954} gave the idea its operational form, now called the distributional hypothesis, which holds that words occurring in similar contexts tend to have similar meanings. Word embeddings implement that hypothesis directly. The Word2Vec model we use in this analysis represents each word by the words that appear around it, so two words are similar when they keep similar company \autocite{mikolov2013}, and a word's nearest neighbors indicate how it is used. Reading congressional rhetoric this way matters for election integrity in particular because the term is both strategically ambiguous \autocite{bach2025let} and lexically ambiguous \autocite{kraft2024lexical}. Democrats and Republicans use it at comparable rates, so counting it or scoring it against a fixed dictionary cannot reveal whether they mean the same thing. Only the company the term keeps can make this comparison meaningfully.

Drawing on insights from symbolic politics, framing, and problem definition, we treat election integrity as a flexible, high-valence term, the meaning of which depends on the narratives elites attach to it \autocite{benford2000framing, edelman1964symbolic, gamson1989media}.

\subsection{Research questions and expectations}

In this study we examine how MOCs construct, contest, and repurpose election integrity over time via their e-newsletters. This motivates three research questions:

\begin{description}
  \item[RQ1:] Do Democratic and Republican MOCs use the same election-integrity vocabulary in semantically different ways, and has that divergence grown over time?
  \item[RQ2:] Has within-party usage of election-integrity vocabulary changed over time?
  \item[RQ3:] What words and themes surround each party's use of election-integrity terms, and what do these semantic neighborhoods suggest about how each party frames election integrity?
\end{description}

We posit that the run-up to the 2020 presidential election sparked a change in how American political parties talk about election integrity---namely, that Democrats support increasing voter access while Republicans support decreasing voter access \autocite{bowler2018partisan, coll2022understanding, hasen2012voting}.

\section{Data and Method}
\label{sec:methods}

Our data come from DCInbox \autocite{cormack2017dcinbox}, a continuously updated archive of 205,861 congressional e-newsletters from 2009 onward, from which we constructed a study corpus of 10,420 newsletters (2010--2025).

\subsection{Data filtering and corpus construction}

We applied a three-tier filter to newsletters from Democratic and Republican members (Independents excluded). Table S1 in the Supplementary Material shows each matching rule and the document yield for all three tiers. Tier 1 matched any of 29 core terms supplied by election-integrity subject matter experts on our team and required no similarity score. Tier 2 required both a contextual or administrative keyword and a similarity score above 0.30. Tier 3 captured documents above 0.50 with no keyword match, recovering 925 documents that keyword search alone would have missed. Similarity scores are the maximum cosine similarity between a Sentence-BERT (all-mpnet-base-v2) embedding of each newsletter’s subject and body, boilerplate removed. Together the three tiers yield 10,420 documents (5.1\% of the original corpus), spanning January 2010 through September 2025 and representing 52.4\% Republican and 47.6\% Democratic communications. The complete term lists and the query templates appear in Tables S2 and S3.

\subsection{Text processing}

We tokenized documents, converted text to lowercase, and removed English stopwords, punctuation, and tokens shorter than three characters. We detected bigrams and trigrams with Gensim's Phrases (min count 5, threshold 10). Because our target terms are single words whose trajectories we track directly, we then reversed any merge that would absorb one of them: ``voter fraud'', for example, is returned to ``voter'' and ``fraud'' so that the standalone terms remain above the minimum frequency threshold in sparse party-years. Phrases containing no target term, such as ``health care'', are left merged. This applies to ``election integrity'' as well; the Semantic neighborhoods section explains why we treat it as two words rather than one.

\subsection{Diachronic word embeddings}

To measure semantic change over time we follow a standard diachronic word embedding approach \autocite{hamilton2016diachronic, kutuzov2017tracing, kutuzov2018diachronic}, training a separate Word2Vec model for each time period rather than a single model over the pooled corpus. Every model reported in this study uses the same configuration: the continuous skip-gram architecture with 50-dimensional embeddings, a context window of five tokens, a minimum word frequency threshold of five occurrences, 40 training epochs, and negative sampling with five negative examples per positive example. We fit one model per party per year (32 embedding spaces). Training is single-threaded with a fixed seed and hash randomization disabled, so every run is exactly reproducible. We aligned each year's space to a 2010 anchor by orthogonal Procrustes rotation fit on the vocabulary shared across all annual models \autocite{hamilton2016diachronic}. Rotation preserves each space's internal geometry while making spaces directly comparable, so cosine distance between a term's vectors in two years measures its semantic shift. The rotation is applied to every vector in the year's space, not only the shared vocabulary.

\subsection{Partisan semantic divergence and shift analysis}

We aligned Democratic models across time to a Democratic anchor (2010) and Republican models to a Republican anchor (2010), following the same orthogonal Procrustes procedure.

For cross-party divergence, we instead rotate each year's Republican space onto its Democratic space (Procrustes on that year's shared vocabulary) and take the cosine distance between the parties' vectors for each target term. Larger distances mean the parties use the term in different contexts.

Even after alignment, two models trained on different samples place an identically used word some distance apart. We refer to this distance as the noise floor. To indicate how our target election terms compare against the noise floor, for each target term and year we take the 100 shared words closest to it in frequency (excluding the other target terms), compute their aligned distances, and report the term's percentile within that set. A term at the 90th percentile is more cross-party divergent than 90\% of comparably frequent words that year; one near the 50th is indistinguishable from them.

Distance shows how much the parties differ in their usage of target terms, but not what the difference is. To see that, we look at which words are used most like the target terms in each party's newsletters. We measure overlap between the two parties' lists of 10 words with the Jaccard similarity: the number of words on both lists divided by the number of words on either list. It ranges from 0 (no words in common) to 1 (identical lists). Because two models trained on different texts will disagree somewhat even when usage is the same, we need a baseline for how much overlap sampling alone would produce. For each party and each year with enough documents, we randomly split that party's newsletters into two equal halves, train a model on each, and compute the Jaccard similarity between the two halves' lists for each term. We treat cross-party overlap as meaningfully low only when it falls below the lower of the two within-party baselines.

\subsection{Within-party shift and term-level asymmetry}

To compare the parties' overall semantic change, we average each party's year-to-year cosine distance across all target terms and transitions. We supplemented mean comparisons with analysis of shifts in timing, identifying which party shifted more during specific periods (i.e., 2019--2020, 2020--2021, 2024--2025). Additionally we report the volume of each term's usage.

\section{Results}
\label{sec:results}

Our analysis of 10,420 congressional newsletters from 2010--2025 shows Democrats and Republicans use the same election vocabulary to invoke different concepts of election integrity. We measure cross-party divergence and within-party shift over time, then semantic neighborhoods. The semantic neighborhoods reveal the sharpest partisan differences, while the distance and shift measures suggest stability.

\subsection{Cross-party divergence over time}

\begin{figure}[h]
\centering
\includegraphics[width=0.9\linewidth]{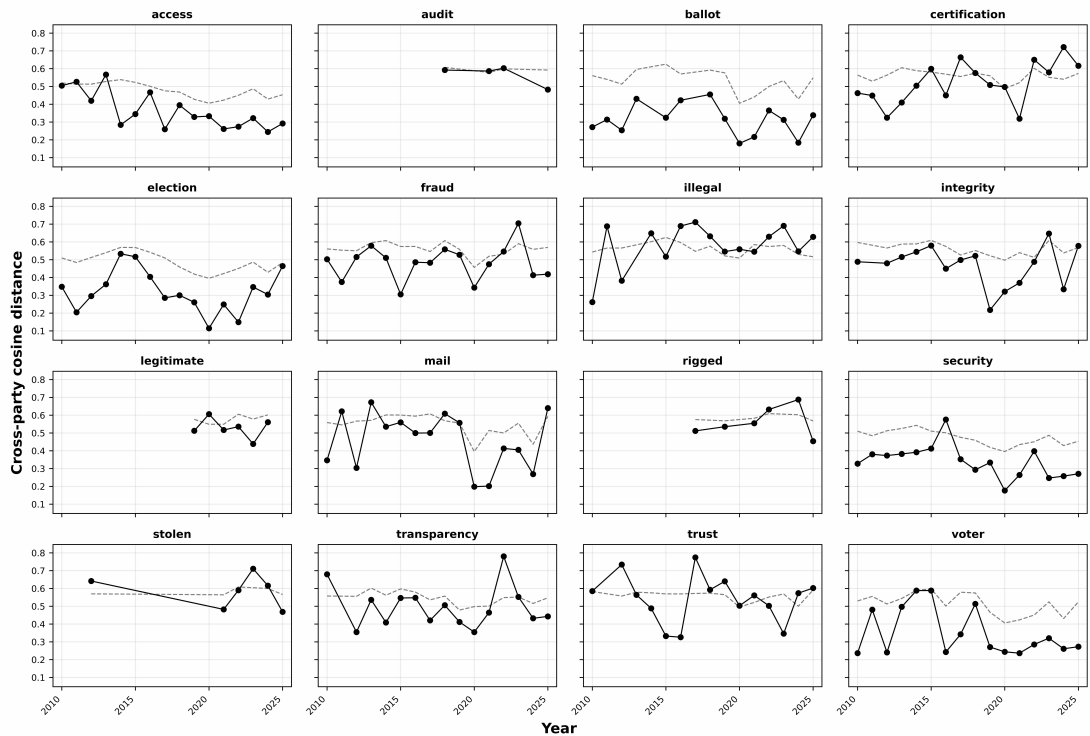}
\caption{\textbf{Cross-party divergence over time.} Each panel shows one target term’s aligned cosine distance between the two parties’ vectors by year.  The dashed line is the noise floor (median aligned distance of the 100 frequency-matched shared words). A year is absent when a term is not mentioned at least five times.}
\label{fig:cross_party_divergence}
\end{figure}

Figure~\ref{fig:cross_party_divergence} tracks partisan divergence over time for our key election integrity terms, and it shows a pattern of fluctuating divergence. It also reveals the emergence of certain terms in political discourse, notably ``stolen'' and ``rigged''. Before 2021, ``stolen'' only cleared the minimum threshold of five mentions per party in 2012, and the term referred to stolen passports, aid, social security numbers and identity theft, and the Stolen Valor Act. In 2021, the term moves directly into election discourse where divergence is nearly at its lowest. Democrats deny the ``big lie'' and Republicans either assert or deny the same claim. As denial of 2020 election results becomes less mainstream, divergence around the term rises. Another emergent term is ``rigged''. In 2017, both parties used the term in reference to a rigged system, economy, and tax code. In 2021, divergence rises as Republicans communicate about a rigged impeachment process.

\begin{figure}[h]
\centering
\includegraphics[width=0.9\linewidth]{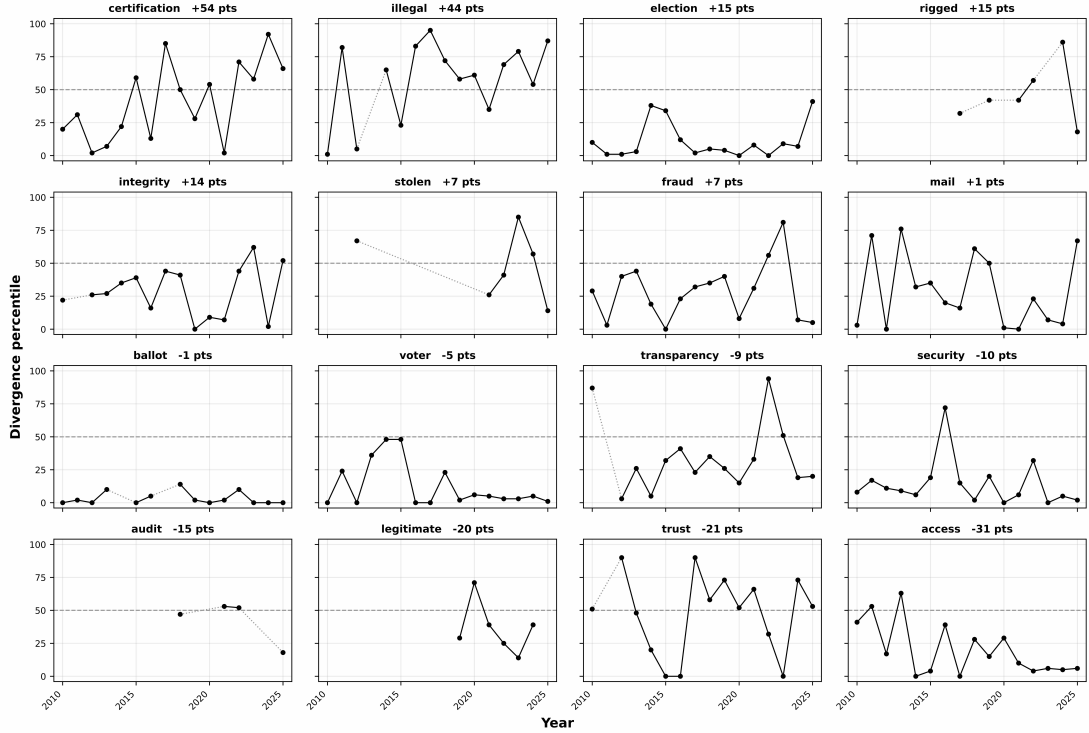}
\caption{\textbf{Cross-party divergence percentile over time.} Each panel shows a term’s divergence percentile among the 100 frequency-matched shared words that year; a term at 90 is more divergent than 90\% of comparable words. Dotted lines indicate a term fell below the five-occurrence minimum. Panels are ordered by the change shown beside each term (mean of last three observed years minus mean of first three).}
\label{fig:cross_party_divergence_percentile}
\end{figure}

Figure~\ref{fig:cross_party_divergence_percentile} further highlights temporal divergence by plotting each term's ranked divergence as a percentile. We also report the difference in percentile points between the last three years a term is observed and the first three years. The term ``certification,'' which appeared in newsletters in all 16 years, rose from the 18th percentile of divergence in its first three years to the 72nd in its last three, a gain of 54 points. Several other terms follow the same upward trajectory, though more modestly: illegal (+44), rigged (+15), integrity (+14), and fraud (+7 points). While ``election'' also shows a 15-point jump, it never rises above the 50th percentile.

\subsection{Within-party shift}

\begin{figure}[h]
\centering
\includegraphics[width=0.9\linewidth]{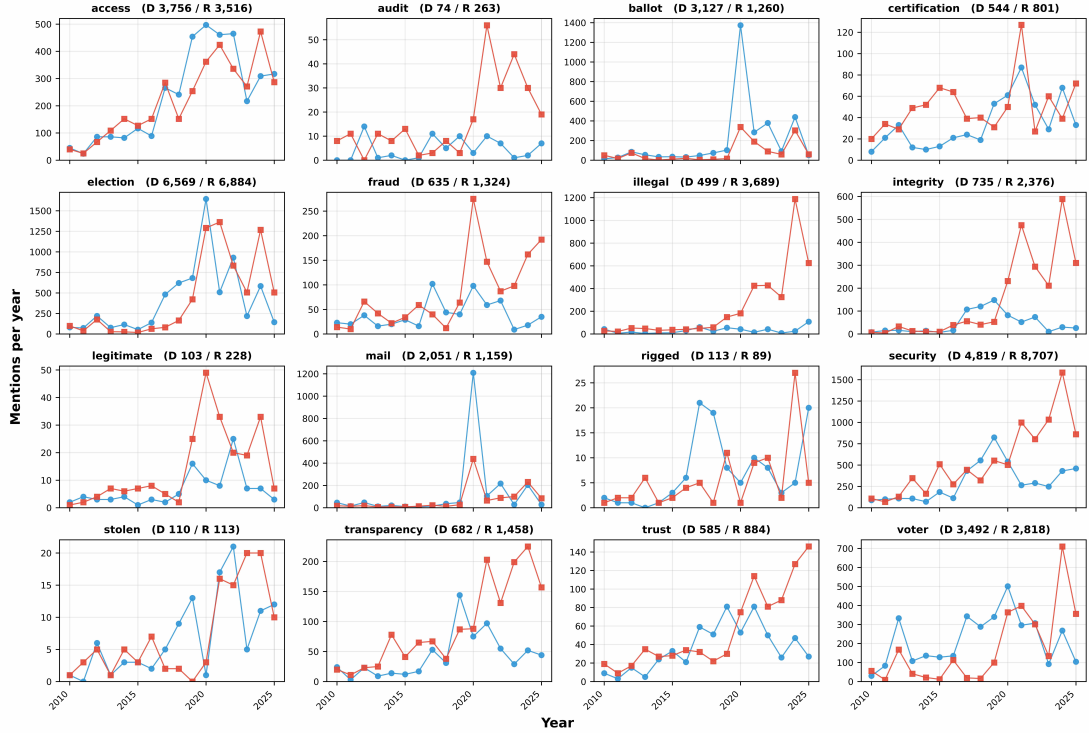}
\caption{ \textbf{Term frequency by party over time.}  Term frequency by party and year (blue, Democratic; red, Republican), with each party’s total shown beside the term.}
\label{fig:term_usage_by_party}
\end{figure}

\begin{figure}[h]
\centering
\includegraphics[width=0.9\linewidth]{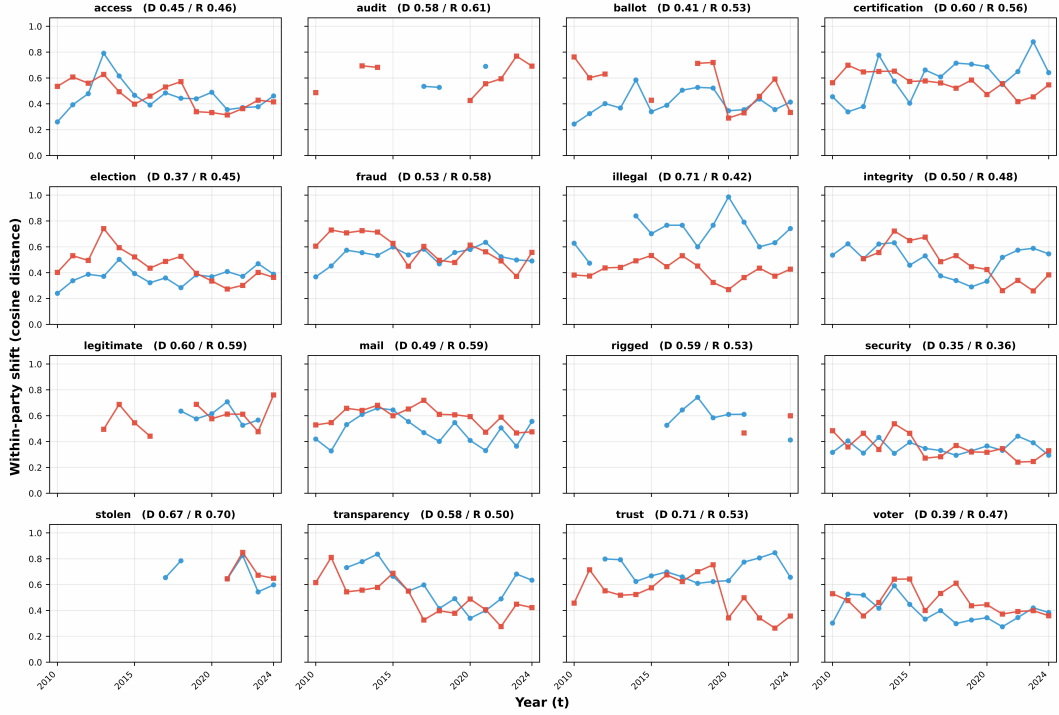}
\caption{\textbf{Within-party semantic shift over time.} Each point is the cosine distance between a party’s vectors for year $t$ and $t+1$ (blue, Democratic; red, Republican); mean shift per party is shown beside the term. }
\label{fig:within_party_shift_over_time}
\end{figure}

The semantic polarization we observe is distributed relatively evenly across parties. Across all 16 target terms, the mean shift in cosine distance for Democrats was 0.53 and 0.52 for Republicans. However, Democrats and Republicans invoke certain terms more heavily and move more substantially on a particular subset of terms, as shown in Figures~\ref{fig:term_usage_by_party} and~\ref{fig:within_party_shift_over_time}.

While it is tempting to look at Figure~\ref{fig:term_usage_by_party} and assume Democrats underwent a huge semantic shift in usage of ``illegal,'' there are two issues with this reading: frequency and sparsity. Republicans use the term ``illegal'' seven times more than Democrats and as a result, their usage is moderately stable in semantic shifts over the 16-year period. Our nearest neighbors analysis confirms this consistency. Over the entire 16-year period, the top K-nearest neighbor to ``illegal'' is either ``immigrant,'' ``immigration,'' or ``alien.'' Democrats, by contrast, displayed a dramatic 0.71 shift overall, but this is likely due to inconsistent usage year to year.

The term ``trust'' is an inverse case. Unlike ``illegal,'' the frequency of ``trust'' is nearly symmetric across parties. At roughly 30 mentions per year per party, the mean within-party shift (Democrats 0.71, Republicans 0.53) cannot be dismissed as a frequency artifact. The more revealing pattern is temporal. Republican usage stabilized sharply after 2020 (0.36) while Democratic usage continued to move. Nearest neighbors analysis suggests why. By 2022 Republican ``trust'' had consolidated around a single institutional frame: ``weaponized'', ``delegitimize'', ``federal bureaucrats'', and by 2024 ``erode'', ``electoral system'', and ``confidence in elections.'' Democratic usage ranged from electoral confidence to various ethical concerns: ``blind trust'', ``special interests'', ``ethical standards.'' This suggests that Republicans consolidated onto a stable framing, rather than that Democrats drifted semantically.

The timing of that consolidation is not arbitrary. It follows the months surrounding the 2020 election, when Republican officials across local, state, and federal races began coordinating messaging that cast doubt on results and advanced claims of stolen or rigged elections \autocite{Barsky2026, mongrain2023suspicious, parks2023eric, tang2023lake}.

``Election'' and ``voter'' tell a different story. Their usage is roughly balanced (``election'' 6,569 Democratic vs.\ 6,884 Republican mentions; ``voter'' 3,492 vs.\ 2,818), yet Republicans show mean shifts of 0.45 and 0.47 against Democratic values of just 0.37 and 0.39. The nearest neighbors analysis in the next section shows this movement is substantive.

\subsection{Semantic neighborhoods}
\label{sec:unigram-choice}

The quantitative measures we have used above tell us when meaning diverged and how much. The semantic network visualizations tell us what actually changed---how Democrats and Republicans now invoke different conceptual frameworks when using the same election integrity vocabulary.

\begin{figure}[h]
\centering
\includegraphics[width=0.9\linewidth]{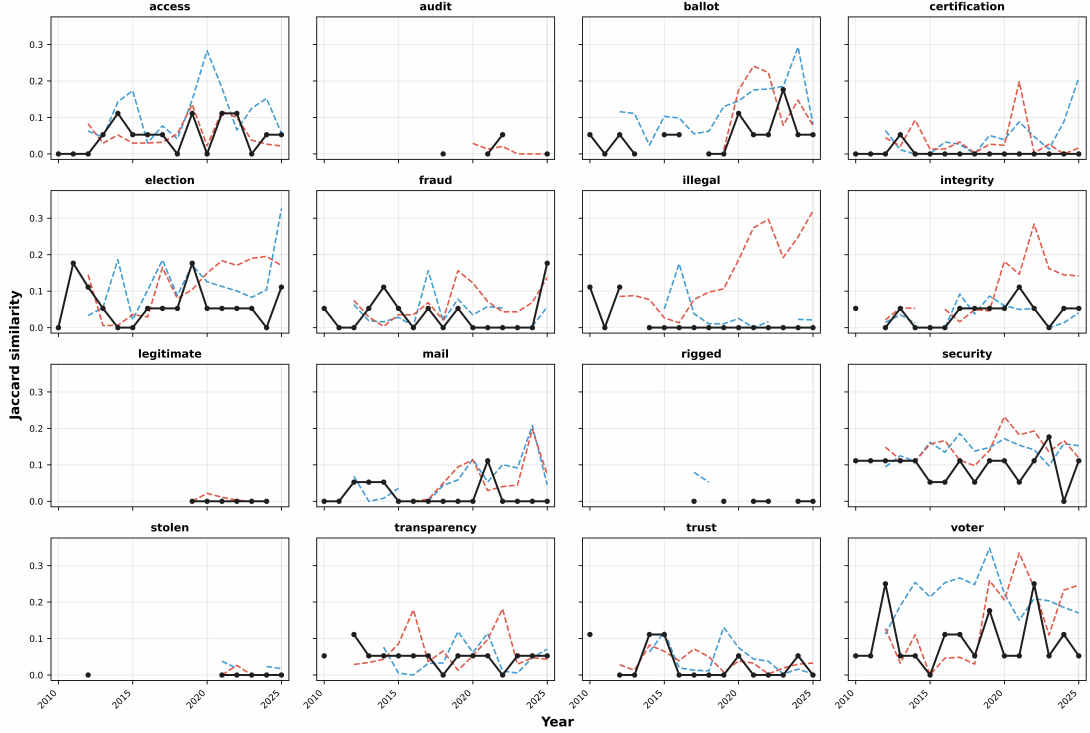}
\caption{\textbf{Semantic neighborhood overlap.} Each panel tracks one target term’s Jaccard similarity (ranges from 0-1, where 1 is identical neighbors) over time across parties and within each individual party. Black: cross-party Jaccard similarity of top-10 neighbor sets. Red and blue: within-party split-half baselines. Cross-party overlap below the lower baseline indicates meaningfully non-overlapping neighborhoods.}
\label{fig:neighbor_overlap_small_multiples}
\end{figure}

The Jaccard similarity between Democratic and Republican semantic neighborhoods for election integrity terms is low across all periods, as shown in Figure~\ref{fig:neighbor_overlap_small_multiples}. Democrats and Republicans are using the same terms to invoke non-overlapping conceptual networks.

Overlap is not entirely absent, and where it appears it is informative. In 2020 the shared neighbors of ``ballot'' are ``provisional'' and ``absentee''---administrative vocabulary both parties adopted when adjusting election procedures for the pandemic. This is consistent with bipartisan efforts to expand the means by which voters could participate that year \autocite{persily2021miracle}. ``Registration'' is a shared neighbor of ``voter'' in 14 of the 16 years we observe, and in 2012 the parties also shared ``laws'', ``identification'', and ``verify''. Parties share a common administrative vocabulary but differ in the language they use to build around it.

\begin{figure}[h]
    \centering
    \begin{subfigure}{0.9\linewidth}
        \centering
        \includegraphics[width=\linewidth,height=0.38\textheight,keepaspectratio]{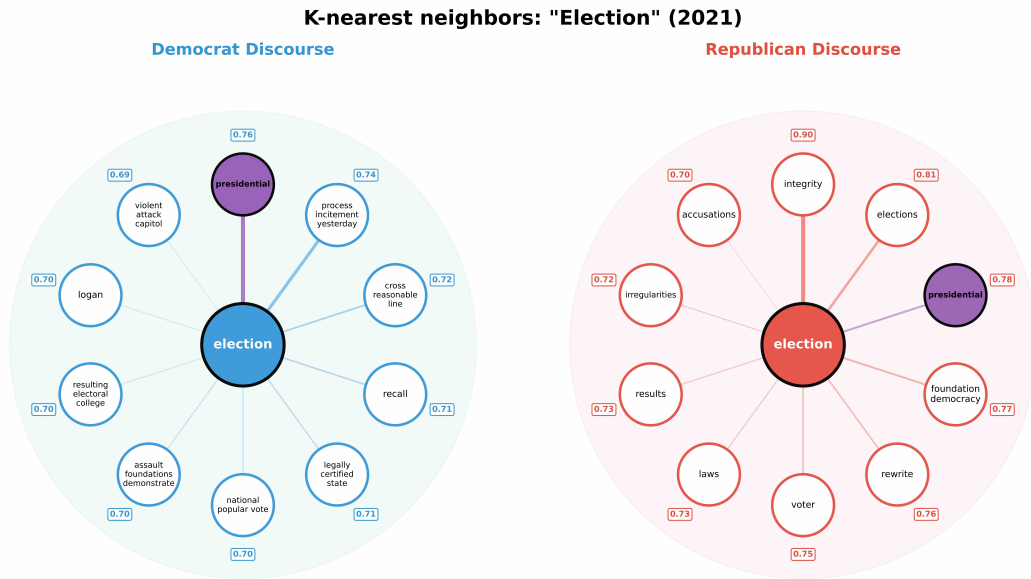}
        \caption{``election''}
        \label{fig:partisan_election_2021}
    \end{subfigure}
    \hfill
    \begin{subfigure}{0.9\linewidth}
        \centering
        \includegraphics[width=\linewidth,height=0.38\textheight,keepaspectratio]{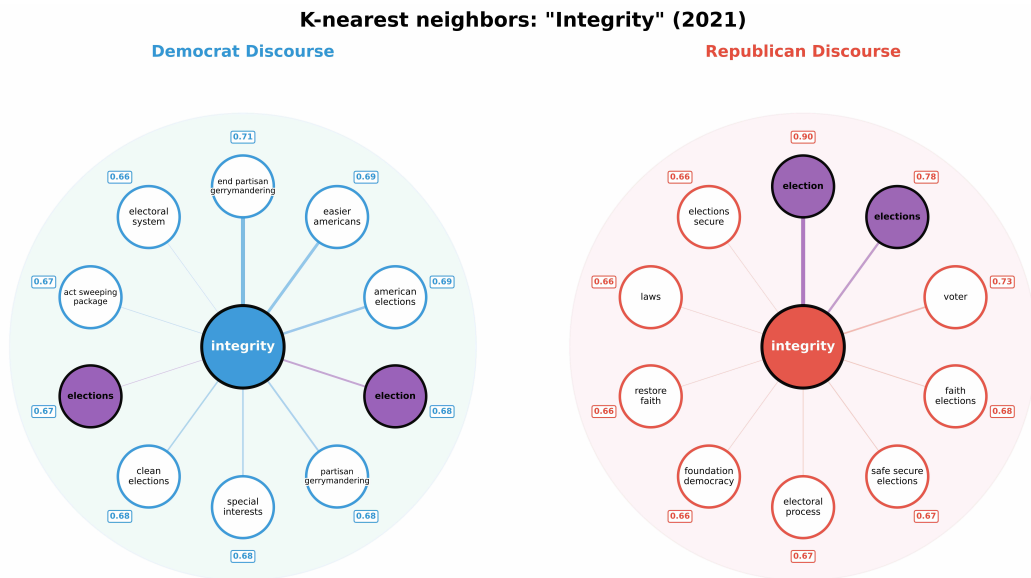}
        \caption{``integrity''}
        \label{fig:partisan_integrity_2021}
    \end{subfigure}
    \caption{\textbf{Top 10 nearest semantic neighbors in 2021 for (a) “election” and (b) “integrity”.} The words each party uses most like the target term. Similarity decreases clockwise; line thickness indicates closeness. Purple marks neighbors shared by both parties.}
    \label{fig:partisan_election_and_integrity_2021}
\end{figure}

Figure~\ref{fig:partisan_election_and_integrity_2021} presents semantic networks for ``election'' and ``integrity'' in 2021. Each network shows the 10 nearest semantic neighbors, the words that most frequently appear in similar context to the target terms in each party's newsletters.

For Democrats, ``election'' sits closest to language about January 6, including ``process incitement yesterday'', ``cross reasonable line'', and ``violent capitol attack''. In contrast, Republicans completely ignore the events of January 6 in their semantic neighborhood. They connect ``election'' to ``integrity'' and vice versa, but at the same time question the election's legitimacy: ``results'', ``irregularities'', and ``accusations''. We see a similar pattern in the ``integrity'' network. While both parties connect ``integrity'' with ``election'', Democrats connect ``integrity'' to ``partisan gerrymandering'', ``special interests'', and ``clean elections'' while Republicans again invoke ``laws'', ``electoral process'', and ``safe secure elections''. The timing of the legislative vocabulary on the Republican side is not incidental: In 2021 several Republican-led states reviewed and passed more restrictive voting laws, and created or reinvigorated dedicated election integrity units within state government \autocite{clark2024policing, greenwood2022florida, kennedy2022desantis, paxton2021integrity, ura2021abbott}.

The parties also differ in the phrases they use to contextualize election integrity. In 2021, 64\% of Republican uses of ``integrity'' fall inside the fixed phrase ``election integrity'', compared to 17\% for Democrats, who more often write ``the integrity of our elections''. This is why Republican ``integrity'' sits closer to the singular ``election'' (0.90) than to the plural ``elections'' (0.78), while for Democrats the two forms are effectively tied (0.68 and 0.67). Combined with their contrasting semantic neighborhoods, this supports the idea that the parties are not aligned on the concept of election integrity. This is also why we report ``election'' and ``integrity'' as separate words. Retaining ``election integrity'' as a single, fixed expression would fail to capture both party constructions.

\begin{figure}[h]
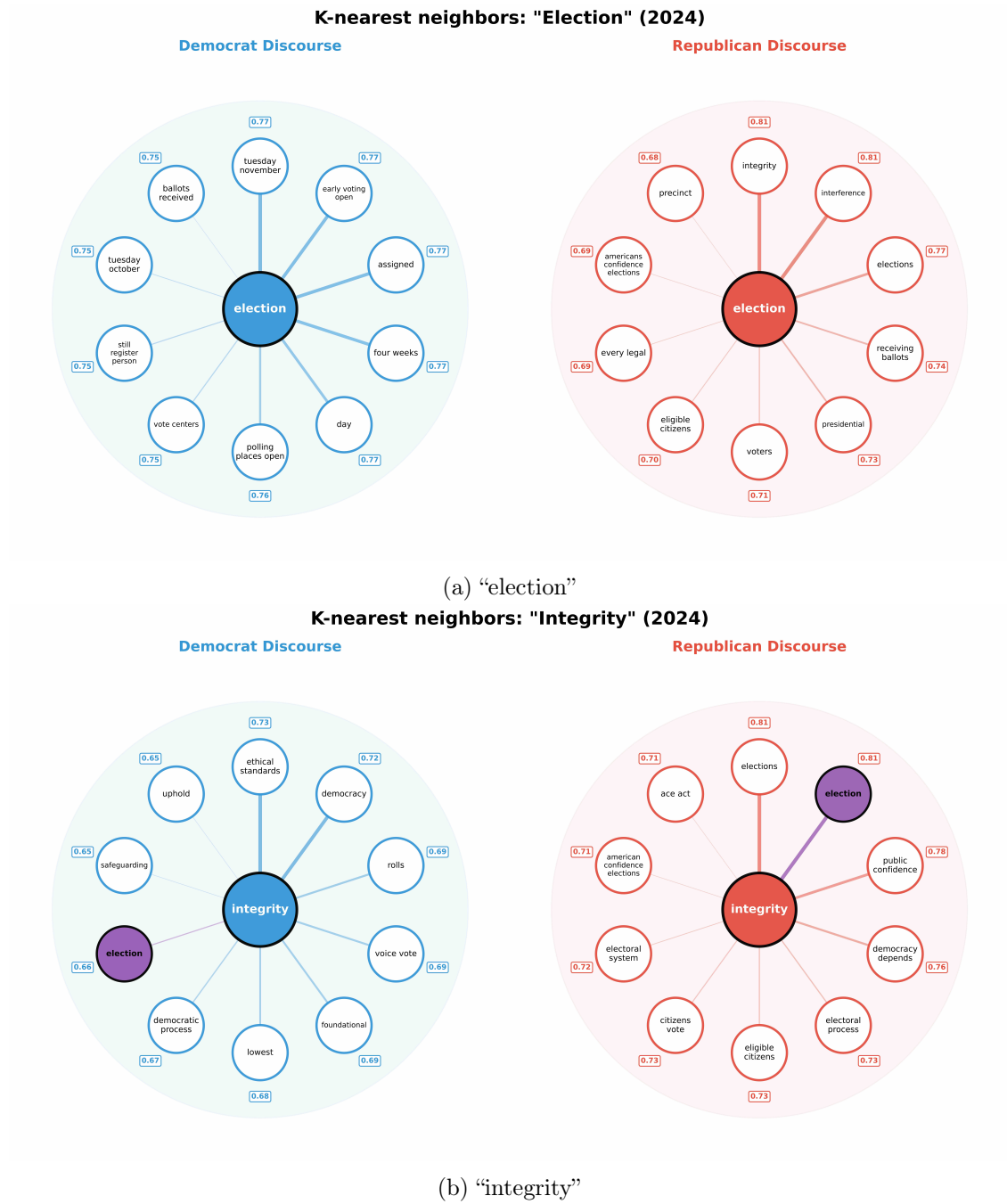

    \centering
    \begin{subfigure}{0.9\linewidth}
        \centering
        \includegraphics[width=\linewidth,height=0.38\textheight,keepaspectratio]{Figures/partisan_election_2024.jpeg}
        \caption{``election''}
        \label{fig:partisan_election_2024}
    \end{subfigure}
    \hfill
    \begin{subfigure}{0.9\linewidth}
        \centering
        \includegraphics[width=\linewidth,height=0.38\textheight,keepaspectratio]{Figures/partisan_integrity_2024.jpeg}
        \caption{``integrity''}
        \label{fig:partisan_integrity_2024}
    \end{subfigure}
    \caption{\textbf{Top 10 nearest semantic neighbors in 2024 for (a) “election” and (b) “integrity”.} }
    \label{fig:partisan_election_and_integrity_2024}
\end{figure}

Figure~\ref{fig:partisan_election_and_integrity_2024} shows the same two networks for 2024. For ``election'', Democrats use language that is tied entirely to voting access: ``tuesday november'', ``early voting open'', ``polling places open''. Republicans connect the term to ``integrity'', but also ``interference''. Notably, Republicans begin to use language tying election integrity to immigration: ``eligible citizens'' and ``every legal''. This pattern continues when looking at ``integrity''. Democrats connect the term to ``ethical standards'' and ``safeguarding'' the ``democratic process''. Republicans tie it to ``public confidence'' and ``democracy depends'', but also ``eligible citizens'', ``citizens vote'', and the ``ace act'', which is a bill that among other provisions ``reduces certain election assistance payments for states and local jurisdictions that allow noncitizens to vote'' \autocite{uscongress2023hr4563ace}. By 2024 the Republican framing explicitly fuses election integrity with immigration enforcement and citizenship verification, a turn that has been documented in Republican messaging more broadly \autocite{berlinski2023effects, bowler2018partisan, sneed2024purge}.

\begin{figure}[h]
\centering
\includegraphics[width=0.9\linewidth]{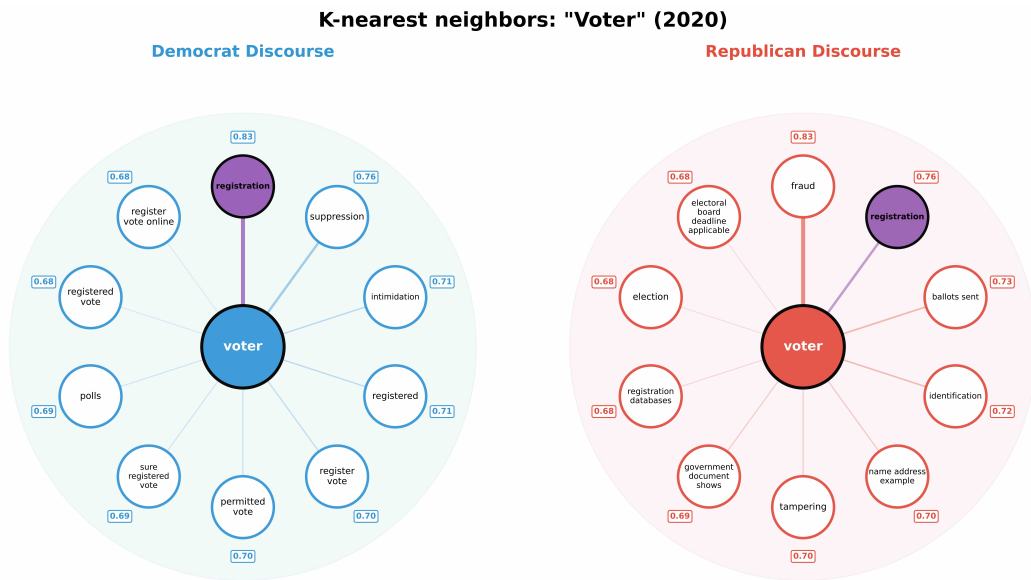}
\caption{\textbf{Top 10 nearest semantic neighbors in 2020 for “voter”. }}
\label{fig:partisan_voter_2020}
\end{figure}

The pattern extends to other terms as well. Consider the semantic neighborhood around ``voter'' in Figure~\ref{fig:partisan_voter_2020}. In 2020, Democrats connected the term almost entirely to ``registration'', only straying from the topic when talking about ``suppression'' and ``intimidation''. Republicans prioritized ``fraud'' above ``registration'' and also used ``identification'' and ``tampering''.

\begin{figure}[h]
\centering
\includegraphics[width=0.9\linewidth]{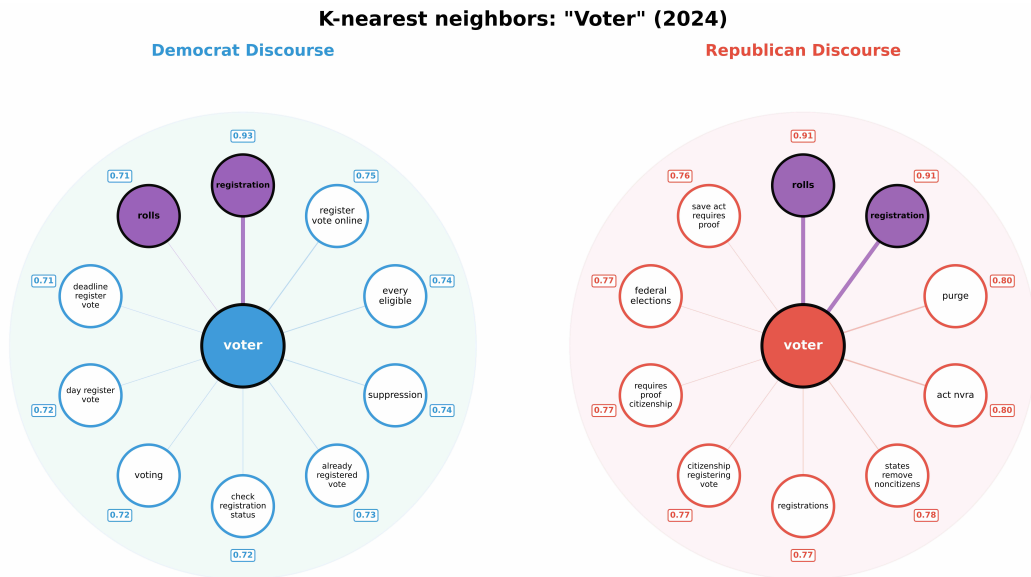}
\caption{\textbf{Top 10 nearest semantic neighbors in 2024 for “voter”. }}
\label{fig:partisan_voter_2024}
\end{figure}

In 2024, Democrats use procedural language, helping voters navigate the voting process as we see in Figure~\ref{fig:partisan_voter_2024}. Most striking is the Democrats' connection of ``voting'' to ``every eligible''. Republicans focus heavily on verification and restrictive language: ``purge'', ``states remove noncitizens'', ``requires proof citizenship'', ``save act requires proof''.

\section{Discussion}
\label{sec:discussion}

We developed a natural language processing pipeline that analyzes 10,420 congressional e-newsletters from 2010 to 2025 to examine whether election integrity is a contested concept in congressional communication. Distance and shift measures alone would suggest stability: the parties moved at nearly identical rates, and most terms were no more divergent than words of similar frequency. The semantic neighborhoods tell a different story. In one usage, election integrity means securing a threatened process. In the other, it is a matter of expanding access. That contest likely played out across many modes of communication, but we observe it in one official channel. The First Continental Congress gave members the franking privilege so they could inform their constituents, and e-newsletters carry that tradition forward. The channel built to inform now carries two competing definitions of one concept no one opposes.

These findings support treating ``election integrity'' as what \textcite{edelman1964symbolic} would recognize as a political symbol: emotionally resonant, normatively unimpeachable, and ambiguous enough to carry opposing agendas without either side ceding the moral high ground. If meaning is use, as \textcite{wittgenstein1968} and \textcite{firth1957synopsis} held, and the term itself is shared, then its partisan meaning travels through the words that surround it. This makes the polarization of election language easy to miss and hard to contest. Constituents who encounter ``election integrity'' as a symbol in their representative's newsletter may be mobilized to respond to a specific story, one that is largely dependent on the party. Instead of a neutral civic commitment to elections and democracy, they are encouraged to believe either that the integrity of elections is under attack or that their access to the democratic process is in peril. Because these messages arrive through official, publicly funded channels framed as governance rather than campaigning, they carry institutional authority that partisan appeals do not.

These results also inform how we understand the construction of symbolic issues. Symbolic issues leverage causal stories to define political problems and create shared identities around problem origins and their potential solutions. The usual expectation is that such stories are carried by a distinct vocabulary. Our results suggest otherwise. The parties tell their stories with the same words, and the stories differ in the company those words keep. They even differ in how they name the problem: Republicans favor ``election integrity,'' Democrats ``the integrity of our elections.''

The neighborhoods of ``ballot'' show the same split. In 2021, Democrats placed it among ``prepaid,'' ``postage,'' ``postmarked,'' and ``box,'' while Republicans placed it among ``harvesting,'' ``signature verification,'' and ``identification.'' These neighborhoods are consistent with a coded language that in-group members might recognize and align with. This suggests that the same words can create opposing frames, with the difference lying in the company they keep.

The consequences extend beyond rhetoric. When the same phrase justifies expanding access in one context and restricting eligibility in another, it means that constituents are hearing two different stories about which policies appear necessary, which administrative practices appear suspect, and which voters appear legitimate. Election officials, who must administer a single system under both definitions, inherit the tension. Understanding how elites redefine ``election integrity'' is not only a question of political communication but also one of democratic governance: the meaning of election integrity is now being contested in the very language used to defend it.

\subsection{Limitations}
\label{sec:limitations}

Word embeddings trained on sparse data are unstable, and our minimum frequency threshold is low. Every model uses 50 dimensions, and a term needs only five occurrences in a party-year to enter the analysis, so low-frequency terms have little context to constrain their vectors and can appear to diverge or shift for reasons unrelated to meaning. “Illegal” shows the problem most clearly. We treat estimates for low-frequency term-years as lower-confidence, and note that the neighborhood findings the paper rests on come from ``election,'' ``voter,'' and ``integrity,'' the highest-frequency terms in the corpus.

Our within-party shift values are not benchmarked against a noise floor. Two models trained on different years place an identically used word some distance apart, so the absolute year-to-year distances we report (around 0.5) are inflated by retraining. We correct for this in the cross-party analysis with frequency-matched percentiles but not for within-party shift, so we interpret within-party shifts only in relative terms, comparing parties and periods.

We held the model configuration fixed. We chose standard settings from the diachronic embedding literature we follow 
\autocite{hamilton2016diachronic,kutuzov2018diachronic} and trained with a fixed seed so every result is reproducible. A sensitivity analysis across seeds, dimensionality, context window, and anchor year would show how far the term-level estimates move under alternative specifications.
Our filtering reduced but did not eliminate non-electoral uses of the target terms. Democratic neighborhoods for ``trust'' include ``public land'' and ``blind,'' and ``stolen'' and ``integrity'' show similar residue. Because Word2Vec assigns one vector per word, these senses blend into the electoral one, and some measured divergence may reflect a difference in topic rather than framing. The semantic queries used to build the corpus (Table S3) also lean toward fraud and security framings over access framings, which may have admitted more Republican-style content at Tiers 2 and 3.

E-newsletters describe elite framing, not public reception. Our claims about how constituents might read these messages are interpretations grounded in the symbolic politics and framing literatures, not measured effects.

Finally, we studied two parties on one channel. We excluded Independents and analyzed only e-newsletters, which reach an audience that opted in to receiving them. Elite rhetoric is often more polarized in partisan-friendly venues than in institutional ones \autocite{green2024cross}, so our divergence estimates are likely conservative. Applying the pipeline to floor speeches, press releases, and social media would show whether the same partisan neighborhoods appear where members speak to broader audiences.

\section{Conclusion}
\label{sec:conclusion}

Our findings suggest that members of Congress (MOCs) do not need different words to tell different stories about American elections: they need only to place the same words in different company. Across 10,420 e-newsletters sent between 2010 and 2025, the two parties drew on a largely shared election vocabulary. With a few notable exceptions, most election vocabulary is no more divergent than other ordinary words used at roughly the same frequency, and neither party's language shifted more than the other's. This makes congressional election rhetoric appear stable and largely bipartisan.

But semantic neighborhoods reveal a deeper contest over meaning that distance metrics cannot: the parties are not disagreeing about the same concept so much as describing two different ones. For instance, ``certification,'' originally one of the least contested terms, rose in its divergence 54 percentile points during the study period. ``Rigged'' and ``stolen'' entered election discourse after 2020, and partisan divergence around them rose as they settled there.

By 2024, Democrats surrounded ``voter,'' ``election,'' and ``integrity'' with the language of access and administration. Republicans surrounded the same terms with citizenship verification and border security, fusing election integrity with immigration enforcement. For one party, integrity means protecting eligible voters' ability to cast a ballot. For the other, it means protecting the electoral process from illegitimate ballots. A shared administrative core remains: ``registration'' appears as a common neighbor of ``voter'' in 14 of 16 years. But each party has built a distinct conceptual structure on top of that common foundation.

\section*{Acknowledgments}

The authors would like to thank the reviewers for their thoughtful feedback. We would also like to acknowledge the DCInbox project and Lindsey Cormack for curating the original dataset that this work used for its analysis. Finally, the authors thank Emma M. White for her research assistance. We would also like to thank Kate Mays for early discussions about this work. 

\printbibliography
\clearpage
\appendix
\section*{Supplementary Material}
\label{app:supplementary}
\addcontentsline{toc}{section}{Supplementary Material}

\setcounter{table}{0}
\renewcommand{\thetable}{S\arabic{table}}

\begin{table}[h]
\centering
\small
\caption{\textbf{Multi-tier filter used to construct the study corpus.} Tiers are
applied in order and are mutually exclusive: a document is assigned to the first
tier whose rule it satisfies. Keyword matching is case-insensitive and
unanchored, so a listed term matches within a longer word or phrase (e.g.,
``certify'' matches ``certifying''). Full term lists appear in
Table~\ref{tab:filter-lexicon}; query templates in Table~\ref{tab:semantic-queries}.
Tier counts sum to 10,423; three duplicates were removed, giving the final 10,420.
Percentages in the final column are shares of the 10,420-document filtered corpus.}
\label{tab:filter-tiers}
\begin{tabularx}{\linewidth}{@{}lXcr@{}}
\toprule
Tier & Match rule & Semantic & Documents \\
\midrule
1 & Mentions any of 29 core election integrity terms & --- & 3,451 (33.1\%) \\
2 & Mentions any of 41 contextual or administrative terms \emph{and} clears the similarity threshold & $>0.3$ & 6,047 (58.0\%) \\
3 & Clears the high similarity threshold, no keyword required & $>0.5$ & \phantom{0}925 \phantom{0}(8.9\%) \\
\midrule
\multicolumn{3}{@{}l}{Total} & 10,420 (100\%) \\
\bottomrule
\end{tabularx}
\end{table}

\begin{table}[h]
\centering
\small
\caption{\textbf{Filter term lexicon.} Terms were supplied by election integrity
subject matter experts on the research team. All Tier~2 terms are matched as a
single pooled set.}
\label{tab:filter-lexicon}
\begin{tabular}{@{}p{0.47\linewidth}p{0.47\linewidth}@{}}
\toprule
\multicolumn{2}{@{}l}{\textit{Tier 1 (29 terms)}} \\
\midrule
election integrity & electoral integrity \\
election fraud & electoral fraud \\
election crimes & election integrity unit \\
voter fraud & voter integrity \\
vote integrity & vote accuracy \\
election security & illegal voting \\
illegal vote & non-citizen voting \\
non-citizens vote & voter id \\
voter identification & voter photo id \\
photo id law & photo id requirement \\
voter suppression & voter suppress \\
voter discrimination & fair elections \\
restoring faith in our elections & restoring faith in democracy \\
securing the vote & for the people act \\
voting rights advancement act & \\
\addlinespace[6pt]
\midrule
\multicolumn{2}{@{}l}{\textit{Tier 2 (41 terms)}} \\
\midrule
rigged & stop the steal \\
big lie & election interference \\
election subversion & collusion \\
russian interference & russian meddling \\
russia investigation & foreign interference \\
certification & certify \\
electoral count & redistricting \\
gerrymandering & shelby decision \\
voting rights act & freedom to vote act \\
john lewis voting rights & voter empowerment act \\
save act & 2010 election \\
mail ballot & mail-in ballot \\
mail-in voting & absentee ballot \\
early ballot & vote by mail \\
ballot security & ballot harvesting \\
ballot drop box & voting system \\
dominion voting & es\&s \\
hand count & touchscreens \\
DRE\textsuperscript{a} & voter registration \\
voter roll & vote centers \\
election worker & \\
\bottomrule
\end{tabular}
\\[4pt]
\begin{minipage}{\linewidth}
\footnotesize
\textsuperscript{a} Direct-recording electronic voting machine. Unlike every
other term, ``DRE'' is matched case-sensitively and on word boundaries, so
``DRE'' and ``DREs'' match but ``hundred'' and ``address'' do not.
\end{minipage}
\end{table}

\begin{table}[h]
\centering
\small
\caption{\textbf{Semantic query templates.} Each document was encoded with
Sentence-BERT (all-mpnet-base-v2) and scored by its maximum cosine similarity
across these 19 queries.}
\label{tab:semantic-queries}
\begin{tabularx}{\linewidth}{@{}rX@{}}
\toprule
\# & Query \\
\midrule
1 & concerns about election integrity and voter fraud \\
2 & securing elections and preventing illegal voting and ballot harvesting \\
3 & election security and ballot verification \\
4 & voter identification and registration fraud \\
5 & noncitizen voting and proof-of-citizenship requirements to register and vote \\
6 & rigged elections and stolen votes \\
7 & election administration and certification processes \\
8 & mail-in ballot security, absentee voting, and ballot postmark deadlines \\
9 & voting machines and election technology security \\
10 & election denialism and false fraud claims \\
11 & partisan election rhetoric and delegitimization \\
12 & voter suppression and election access \\
13 & ballot harvesting and election crimes \\
14 & redrawing district lines to gain partisan political advantage \\
15 & redistricting and gerrymandering to dilute minority voting power \\
16 & foreign interference and Russian meddling in U.S. elections \\
17 & protecting elections from foreign influence and disinformation \\
18 & government units with police powers investigating election crimes \\
19 & claims that elections were stolen through widespread coordinated fraud \\
\bottomrule
\end{tabularx}
\end{table}
\end{document}